# A Post-Training Enhanced Optimization Approach for Small Language Models

**Keke Zhai**
zhaikeke@buaa.edu.cn

**Abstract**

This paper delves into the continuous post-training optimization methods for small language models, and proposes a continuous post-training alignment data construction method for small language models. The core of this method is based on the data guidance of large models, optimizing the diversity and accuracy of alignment data. In addition, to verify the effectiveness of the methods in this paper, we used Qwen2-0.5B-Instruct model as the baseline model for small language models, using the alignment dataset constructed by our proposed method, we trained and compared several groups of experiments, including SFT (Supervised Fine Tuning) post-training experiment and KTO (Kahneman Tversky optimization) post-training experiment, as well as SFT-KTO two-stage post-training experiment and model weight fusion experiment. Finally, we evaluated and analyzed the performance of post-training models, and confirmed that the continuous post-training optimization method proposed by us can significantly improve the performance of small language models.

## 1 Introduction

With the rapid development of artificial intelligence technology, the performance of large language models (LLMs) [1] has made significant progress in the past few years. The number of parameters in these models has increased from the initial billion level, such as GPT-2 [2] with 1.5 billion parameters, to today's trillion level, and the scale of training data has also increased from a million level to tens of trillions level of tokens, such as the 15T tokens data of LLaMA 3.1 [3]. The driving force behind this progress are predictable scaling laws [4], which indicate that the performance of the model is directly proportional to its scale. However, with the growth of the model scale, the cost and delay of reasoning have also increased, which limits their applications in resource-constrained environments such as edge devices.

To solve this problem, Small Language Models (SLMs) come into being. For example, Microsoft's Phi-mini series [5], Google's Gemma 2 series [6], Meta's LLaMA 3.1-1B/3B [3], and Alibaba's Qwen-0.5B/1.5B/3B [7], etc. The strategy of large-scale data pre-training small models is adopted. The core idea of these models is to improve performance by increasing the scale of data while keeping the model parameters fixed, so as to maintain lower reasoning costs and delay, and to achieve performance close to large models.

Although small language models have been widely optimized during the pre-training phase, the optimization of post-training phases is relatively scarce. This paper mainly explores the continuous optimization method of post-training of small language models, and proposes a continuous post-training alignment data construction method for small language models. The core of this method is based on the data guidance of large models, optimizing the diversity and accuracy of alignment data, and taking into account the generation safety of the model. In addition, this paper also compares the effects of SFT (Supervised Fine-Tuning) [8] and KTO (Kahneman-Tversky Optimisation) [9] on model performance, and verifies the significant effect of continuous post-training on improving the performance of small language models through continuous post-training optimization based on open-source models.

## 2  Related Work

### 2.1  Small Language Model (SLMs)

Although large models have achieved significant achievements in performance. However, their high computing costs and resource consumption have limited their applications in resource-constrained environments. Small Language Models (SLMs) have attracted more and more attention due to their advantages in resources efficiency and actual costs, as these models demonstrate capabilities comparable to Large Language Models (LLMs) while maintaining a relatively small model size. For example, Microsoft's Phi-2/3 series models [10] have broken existing scaling laws, showing that high-quality data itself is sufficient to build models that can compete with larger models, and Qwen-0.5B/1.5B [7], LLaMA 3.2-1B [3], Tinyllama [11], MiniCPM [12], Shakti [13], etc. These small language models are based on the Transformer decoder model and focus on exploring the training of small models with a much larger number of training tokens than what the scaling laws suggest [14], so that they have made significant breakthroughs in both performance and efficiency, and they have shown huge potential in various application scenarios.

### 2.2  Post-Training Learning

The training of large models can be divided into two stages of pre-training and post-training. Pre-training usually requires a large amount of computing resources and data. The basic model generated by pre-training has the ability to commonly understand human knowledge. The post-training phase is based on the pre-trained base model, through further aligns the model with instructions to follow the ability and better stimulate the various comprehensive abilities of the model. Post-training includes supervised fine-tuning (SFT) [8] and RLHF [15]. SFT has also various related methods proposed, such as LORA [16], QLORA [17]. RLHF is a widely used method that depends on learning reinforcement strategies from human feedback. The process of RLHF includes two stages: first, training a reward model (RM) with human preference data, and then using this reward model to guide the reinforcement learning optimization of the policy model (Policy Model). However, RLHF has several significant problems, such as high memory occupation, unstable training, and complex processes. To solve the complexity of RLHF, the DPO [18] method was proposed. DPO simplifies the RLHF process, transforming the training phase of enhanced learning into a binary classification problem, reducing memory consumption and improves training stability. However, DPO cannot make full use of the reward model, and it is only applicable to pair of preference data, and it cannot process a wider feedback type. Further, KTO (Kahneman-Tversky Optimization) [9] was proposed，the core idea of KTO is to leverage the loss aversion characteristic in the process of human decision-making, and to align the output of the model and human values by optimizing the loss function that considers loss aversion. KTO does not require paired preference data, but using a single feedback (such as "good" or "bad" label), which makes data collection easier and low in cost. The KTO method is simple and easy to use and can be applied to any dataset containing positive and negative labels.

At present, the post-training of small language models is more concentrated on the SFT based on the basic model of pre-training, as well as further DPO and other related post-training optimization. The key of SFT/DPO and other methods is the quality of alignment data. In addition to manual annotation to ensure quality, some generation methods have also been proposed. For example, the automation synthesis method of Magpie [19], this method improves LLM's performance by synthetic alignment data, and there is less research on whether continuous post-training for small language models can further significantly improve model performance, which is also the core work of this paper.

## 3  Post-Training Data Construction

### 3.1  Overall Data Pipeline

To effectively improve the diversity and accuracy of post-training data, this article proposes a post-training data pipeline. The overall architecture of the post-training data pipeline is shown in Figure 3-1:

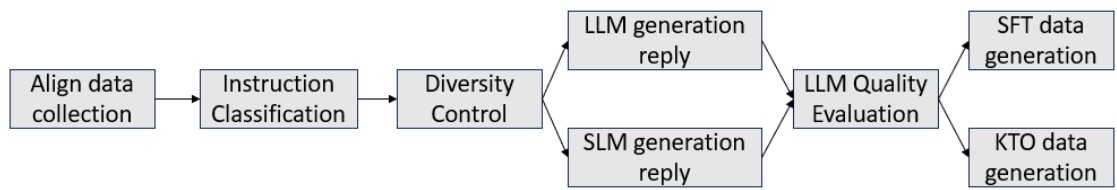

Figure 3-1 Post-Training Data Pipeline

The overall post-processing data pipeline includes alignment data collection, instruction classification, diversity control, large language model and small language model generating response, large language model quality evaluation, and final generation of SFT alignment data and KTO alignment data . The following sections will provide a detailed introduction.

### 3.2  Alignment Data Collection

Our data all comes from the open source datasets on the Internet, mainly including BelleGroup/train_3.5m_CN, BelleGroup/school_math_0.25M, Coig, Camel, Dolly, Alpaca_GPT4 and other open source datasets [20], covering math, code, reasoning and other general instructions. We have standardized the format of these data, and preliminarily classified them based on known types using rule-based methods. According to the content of the task, they are divided into 50 categories. At the same time,performing deduplication of the instruction prompt part,mainly based on the simhash deduplication algorithm, this algorithm can be used for deduplication of massive data, by tokenizing all input instructions to calculate simhash values, establishing indexes, and setting the tolerance threshold k of Hamming distance, where a larger k value results in more deduplication samples, we set k to 3. We ultimately generated approximately 5M records of data, including 4M Chinese instructions and 1M English instructions. The data matching is shown in Figure 3-2:

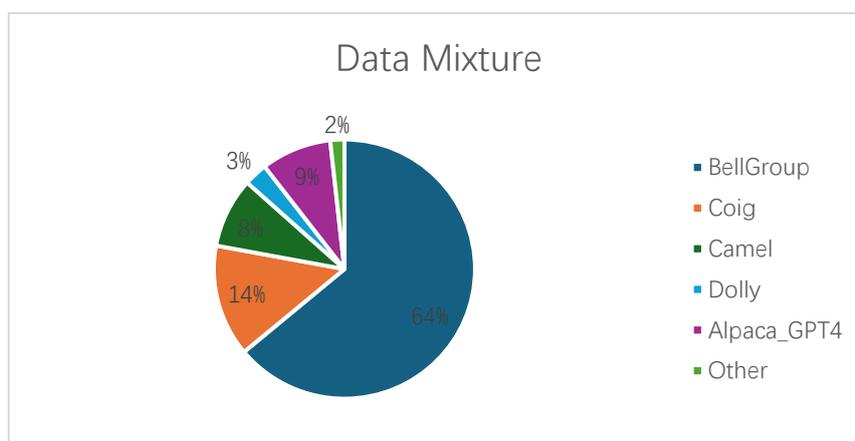

Figure 3-2 Data Matching

### 3.3 Instruction Classification

To ensure the diversity of data, we classified the instructions on the preliminary screened 5M data. The core method is to directly generate a secondary classification label for the prompt part of the instruction through the prompt method based on the large language model. The prompt is designed as follows:

> You are an instruction classification assistant. Please give a secondary classification label to the input instruction text in the format: Primary Category-Secondary Category, output directly without explanation.

Due to resource limitations, we use the Qwen2.5-7B-Instruct model to generate secondary classification label.

### 3.4 Diversity Control

Each sample data has a secondary classification label after instruction classification. To ensure the reliability of the labels, we removed the low-frequency classification labels (the label frequency less than 10, adjustable). We assessed the accuracy of the labels through sampling and removed those samples with relatively low accuracy labels, retaining the 5k+ secondary classification labels, and then ensure that the distribution of classification labels is as diverse as possible by rejecting sampling methods, the number of each classification instruction is relatively balanced, ultimately generating 50w+ diverse samples.

### 3.5 Regenerate Reply

Each instruction of the original open source data has a reply, but to obtain higher quality data, we choose to regenerate the replies, including using a large language model to generate replies, and a small language model to generate replies. Because our optimization baseline for the small language model is Qwen2-0.5B-Instruct, the selection of the large language model is theoretically the larger the better, but it is limited by experimental costs, we select Qwen2.5-7B-Instruct model to generate replies, and directly select Qwen1.5-0.5B-Chat as the small language model to generate replies, and finally each instruction has 3 replies: the original reply, the large language model reply, and the small language model reply.

## 3.6 Quality Evaluation

The quality of post-training alignment data is very important, so the key to quality evaluation is to select samples that are both safe and the highest reply quality from the three replies. To this end, we have adopted a variety of methods: such as the perplexity of the reply content, the large model's safety judgment on the instruction and reply, and the large model's comprehensive scoring on the instruction and reply.

The first method is to use a large language model to calculate the perplexity of each reply content. Perplexity is an indicator to measure the performance of the language model, reflecting the model's predictive ability for text. Specifically, perplexity is calculated by taking the average of the negative logarithm of the probability of each word in the model. Assuming a sentence $s = (w_1, w_2, ..., w_n)$ with a length n, its perplexity is defined as:

$$PP(W) = P(w_1 w_2 ... w_n)^{-\frac{1}{n}} = \sqrt[n]{\frac{1}{P(w_1 w_2 ... w_n)}}$$

Taking the logarithm first and then the exponent, the transformation becomes the following formula:

$$\text{Perplexity}(s) = e^{-\frac{1}{n}\sum_{i=1}^{n} \ln(p(w_i|w_1 w_2 ... w_{i-1}))}$$

In the formula, n is the length of the text sequence, w(i) is the i-th word in the sequence, and $p(w_i|w_{1:i-1})$ is the probability of the i-th word given the previous words. From the formula, it can be seen that the larger the sentence probability, the better the language model, and the smaller the perplexity. The formula above shows that PP(W) essentially becomes a cross-entropy function with an exponential base, so when we want to calculate perplexity, we can directly calculate the cross-entropy, where the size of the base is not important, in this paper, the base is set to 2. For reply content, the smaller the perplexity, the better the reply is often indicated, so perplexity can be used to distinguish the quality of replies.

The second method is to use the large model to make a safety judgment on each instruction and reply, and assign a safety score to each instruction pair. We adopt a 3-point system, including {1, 0.5, 0}, where 1 point represents that both the instruction and the reply are very safe, 0.5 represents that one of the instruction and reply is unsafe, and 0 represents that both the instruction and the reply are unsafe.

The third method is to use the large model to make a comprehensive quality judgment on the replies of each instruction, from the dimensions of whether the intention of the instruction is followed, the accuracy, completeness, and relevance of the reply, etc., assigning a quality score to each reply. We adopt a 3-point system, including {1, 0.5, 0}, where 1 point represents a very good reply, 0.5 represents an average reply, and 0 represents a poor reply.

In the end, we use the perplexity value of the calculated reply, the safety score of the large model, and the quality score of the large model as the features of each sample, which facilitates further sample selection based on these features.

### 3.7 Post-Training Alignment Data Construction

Post-training alignment data generation includes SFT (Supervised Fine-Tuning) alignment data generation and KTO (Kahneman-Tversky Optimization) preference alignment data generation. The core method is to select based on the instruction-reply perplexity values, safety scores, quality scores, and other features built from the quality assessment of the previous stage for each sample. We select the best reply for each sample instruction as the SFT post-training dataset, and on the basis of the best reply quality of the SFT data, we also retain the worst replies to form the KTO preference alignment data. Through comprehensive screening based on deduplication, perplexity, safety scores, quality scores, etc., we ultimately constructed approximately 100,000 SFT datasets and 150,000 KTO datasets, the ratio of positive feedback to negative feedback samples in the KTO dataset is 2:1.

## 4 Experiment and Evaluation

### 4.1 Experimental Methods and Evaluation Indicators

To verify the effectiveness of the methods in this paper, we used Llama-Factory [21] as the training and reasoning framework, and Qwen2-0.5B-Instruct as the baseline model for small language models, and conducted post-training experiments and model weight fusion experiment using the SFT and KTO datasets we constructed.

In terms of training parameters, we focus on adjusting the learning rate and batch size, taking into account the balance between model convergence speed and computational resource consumption. We follow the method of Kaplan et al.[4] to determine the batch size based on expected loss, and select the optimal learning rate based on multiple experimental comparisons.

In addition, to verify the effectiveness of the methods we proposed, and strive to evaluate the model performance in a comprehensive and objective manner. We have selected a widely recognized benchmark test set to ensure the authority and universality of the evaluation. We adopted two industry standard evaluation tools, Qwen-Eval [22] and Human-Eval [23], to conduct a detailed evaluation of the model's performance in multiple dimensions.

The test datasets we have selected include gsm8k, mmlu, cmmlu, ceval, and HumanEval. These datasets cover multiple aspects from basic language understanding to complex reasoning and common sense judgments, enabling a comprehensive examination of the model's performance on different tasks. Through dataset evaluation, we can conduct in-depth analysis and evaluation of the model's generalization ability, reasoning ability, and the degree of mastery of human ommon sense, and verify the effectiveness of our methods. We can also gain a deep understanding of the model's advantages and limitations in handling various language tasks.

### 4.2 SFT Continuous Post-Training Experiment

The baseline model for continuous SFT post-training is Qwen2-0.5B-Instruct. We used the constructed enhanced SFT dataset for continuous SFT alignment learning. The experimental process is shown in Figure 4-1:

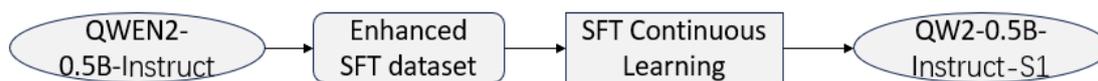

Figure 4-1 Schematic Diagram of Continuous SFT Alignment Learning

Due to the experiment was conducted on a single RTX 3090 card, we set a smaller batch size of 4, with gradient accumulation of 250 and a warmup ratio of 0.005. By trying different learning rates, we found that the learning rate of 5e-8 was more appropriate. It is a relatively small value that can ensure the stability of the model during the fine-tuning process, avoiding excessive update steps to lead to fluctuations in model performance. At the same time, we used the Cosine Annealing Scheduler to adjust the learning rate. This scheduler can gradually reduce the learning rate according to the preset cycle, so as to achieve more detailed optimization in the later stages of training. The Cosine Annealing Scheduler helps the model to explore parameter space more smoothly as it approaches convergence, improving the final performance.

The SFT training loss curve is shown in Figure 4-2:

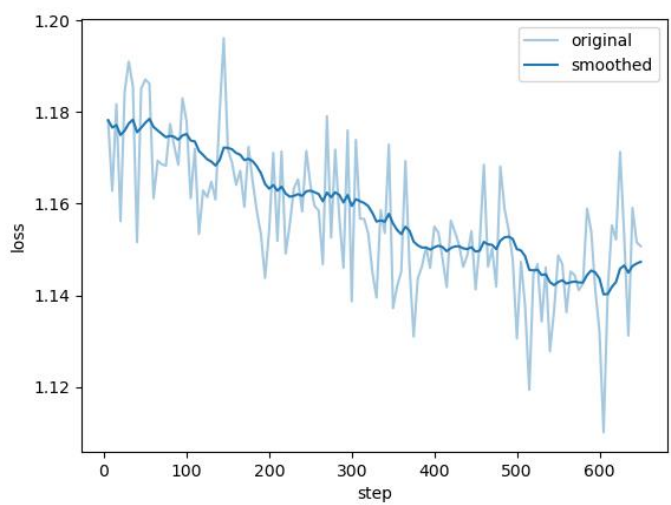

Figure 4-2 SFT Post-Training Loss Curve

The benchmark evaluation results are shown in Table 4-1:

| Evaluation Set | Qwen2-0.5B-Instruct | QW2-0.5B-Instruct-S1 | Gain |
| --- | --- | --- | --- |
| mmlu | 37.9 | 39.07 | +1.17 |
| cmmlu | 45.48 | 45.80 | +0.32 |
| ceval | 45.2 | 49.52 | +4.32 |
| HumanEval | 17.1 | 27.44 | +10.34 |
| gsm8k | 40.1 | 42.00 | +1.9 |

Table 4-1 Benchmark Evaluation Results of Continuous SFT Post-Training

Through the evaluation and analysis of benchmark indicators, we can see that further post-training alignment learning on the enhanced SFT dataset we constructed can significantly improve model performance.

## 4.3 KTO Continuous Post-Training Experiment

To make the experiment comparable, the baseline model selected for continuous KTO post-training is also Qwen2-0.5B-Instruct. We used the constructed enhanced KTO preference dataset for continuous KTO alignment learning. The experimental process is shown in Figure 4-3.

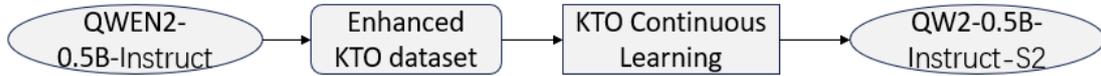

Figure 4-3 Schematic Diagram of Continuous KTO Post-Training

In the KTO experiment, the optimal learning rate was set to 1e-8 through the comparison of multiple experiments. It also set a smaller batch size 4, with gradient accumulation of 250 and a warmup ratio of 0.005, using the Cosine Annealing Scheduler to adjust the learning rate.

The KTO training loss curve is shown in Figure 4-4:

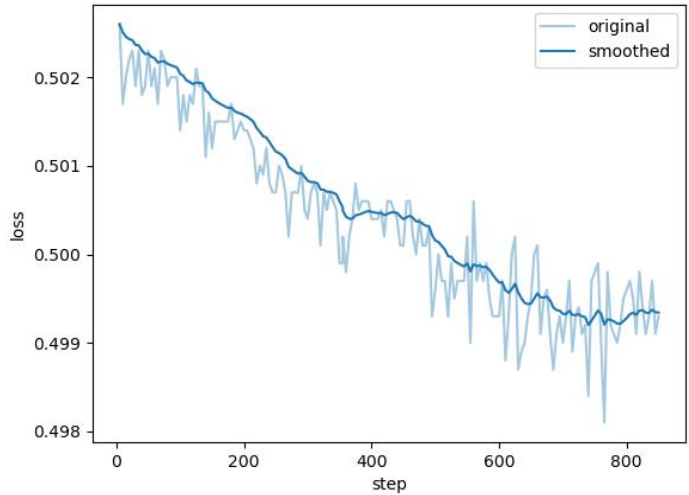

Figure 4-4 KTO Post-Training Loss Curve

The benchmark evaluation results are shown in Table 4-2:

| Evaluation Set | Qwen2-0.5B-Instruct | QW2-0.5B-Instruct-S2 | Gain |
|---|---|---|---|
| mmlu | 37.9 | 38.74 | +0.84 |
| cmmlu | 45.48 | 46.84 | +1.36 |
| ceval | 45.2 | 49.42 | +4.22 |
| HumanEval | 17.1 | 27.44 | +10.34 |
| gsm8k | 40.1 | 41.17 | +1.07 |

Table 4-2 Benchmark Evaluation Results of Continuous KTO Post-Training

Through the evaluation and analysis of benchmark indicators, we can see that further post-training alignment learning on the enhanced KTO dataset we constructed can significantly improve model performance, but it is basically the same as the continuous post-training effect of SFT.

## 4.4 SFT-KTO Two-Stage Experiment

To verify whether continuous SFT alignment training followed by further KTO post-training can continue to improve model performance, we designed a two-stage SFT-KTO experiment, and the experimental process is shown in Figure 4-5.

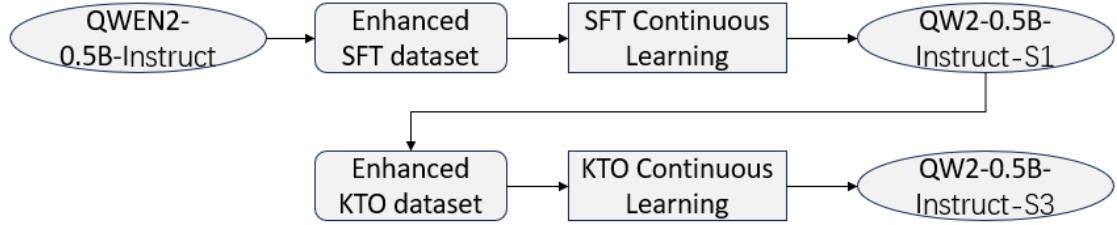

Figure 4-5 Schematic Diagram of Two-Stage SFT-KTO Post-Training Alignment

We selected the convergence point of the model with SFT post-training learning rate of 5e-8 as the QW2-0.5B-Instruct-S1 model for the two-stage training. For the KTO post-training, the optimal learning rate was set to 3e-8 after comparing multiple experiments. The same setting of smaller batch size was 4, with gradient accumulation of 250 and a warmup ratio of 0.005, using the Cosine Annealing Scheduler to adjust the learning rate.

The benchmark evaluation results are shown in Table 4-3:

| Evaluation Set | Qwen2-0.5B-Instruct | QW2-0.5B-Instruct-S3 | S3 Gain | S2 Gain | S1 Gain |
|---|---|---|---|---|---|
| mmlu | 37.9 | 39.32 | +1.42 | +0.84 | +1.17 |
| cmmlu | 45.48 | 47.4 | +1.92 | +1.36 | +0.32 |
| ceval | 45.2 | 50.16 | +4.96 | +4.22 | +4.32 |
| HumanEval | 17.1 | 28.05 | +10.95 | +10.34 | +10.34 |
| gsm8k | 40.1 | 42.5 | +2.4 | +1.07 | +1.9 |

Table 4-3 Benchmark Evaluation Results of Two-Stage SFT-KTO Post-Training

Through the evaluation and analysis of benchmark indicators, we can see that the model performance based on the two-stage experiment can be further improved compared to either continuous SFT or continuous KTO post-training alone, which can further improve the effectiveness of SFT-KTO two-stage continuous post-training for SLM.

## 4.5 Weight Fusion Experiment

Different stages of post-training may have different focuses on improving model performance. Therefore, to make the overall performance improvement of the model balanced, we designed weight fusion experiments to verify the impact of weight fusion on model performance. The weight fusion adopts a strategy of averaging weighted weights. We validated this by setting different weighted coefficients for the three post-trained models of QW2-0.5B-Instruct-S1, QW2-0.5B-Instruct-S2, and QW2-0.5B-Instruct-S3 and then fusing their weights.

The benchmark evaluation results are shown in Table 4-4:

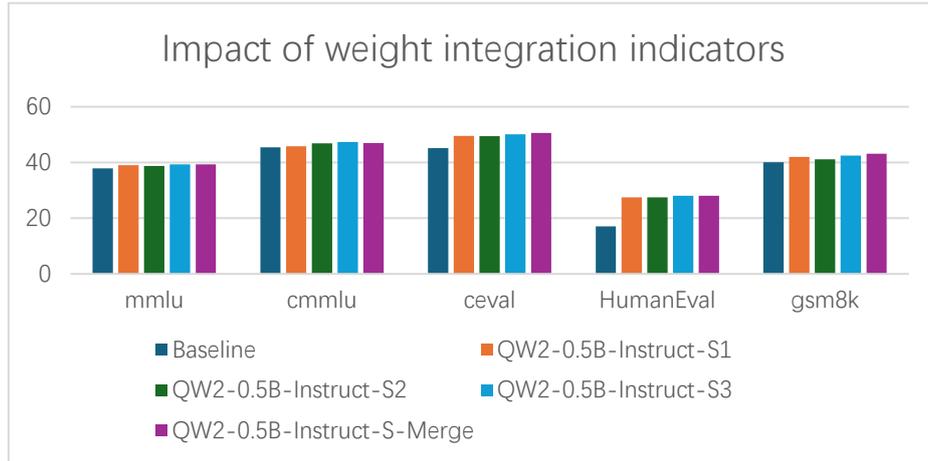

Table 4-4 Benchmark Evaluation Results of Weight Fusion

Through the evaluation and analysis of benchmark indicators, we can see that the overall performance of the model is further enhanced and relatively balanced on each indicator after adopting the weight fusion strategy. This indicates that fusing the weights of models trained continuously with different methods based on QW2-0.5B-Instruct, compared to using a single training method, can effectively improve the overall performance of the model.

### 4.6 The Impact of Learning Rate Hyperparameter

Through SFT and KTO post-training experiments, we found that using different learning rates has varying impacts on the indicators of the trained model. We learned that the learning rate of Qwen2-0.5B-Instruct during the pre-training phase is 1e-6. When we used the same learning rate on 100000 SFT samples, we found that the loss converged quickly, but the overall indicators were poor. It was not until the learning rate reduced by two orders of magnitude to 5e-8 that the overall indicators began to steadily improve, especially the gsm8k indicator has significantly improved above the baseline. Therefore, it is very important to select the appropriate learning rate for continuous post-training. It directly affects the performance of the model.

The loss trend diagrams corresponding to different learning rates are shown in Figure 4-6:

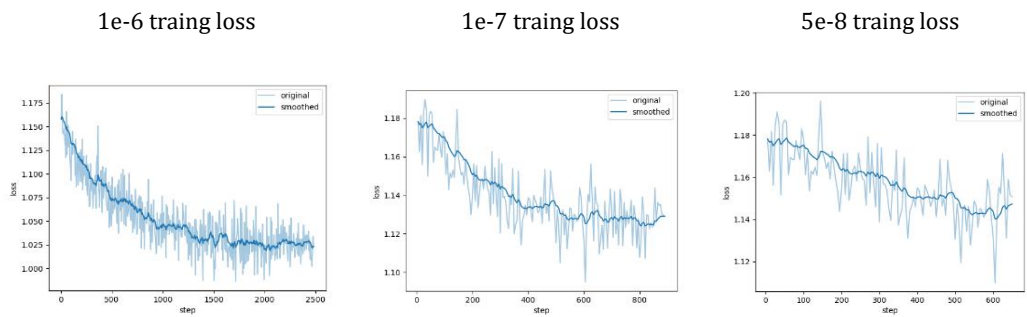

Figure 4-6 SFT Post-Training Loss Trend Diagrams with Different Learning Rates:

In the diagram above, when the learning rate is 1e-6, the batch size is 0.3k. When the learning rates are 1e-7 and 5e-8, we increase the batch size to 1k. After approximately 6 epochs, the model tended to converge. We evaluate the indicators at the convergence points of different learning rates, and the test results are shown in Figure 4-7:

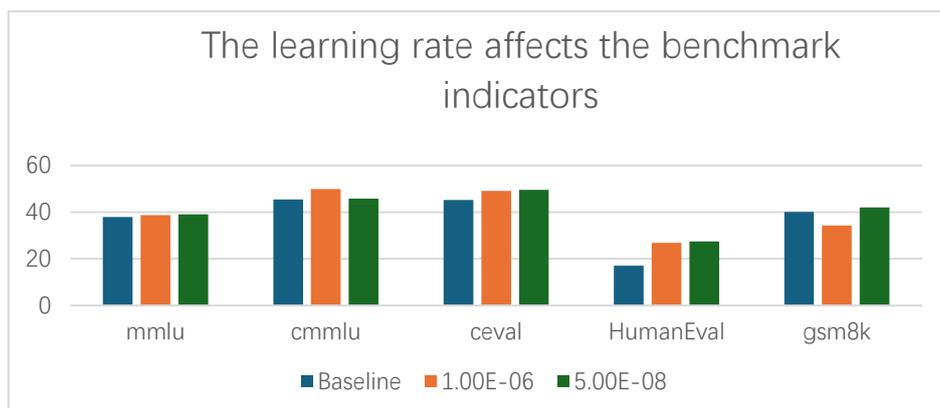

Figure 4-7 Impact of Different Learning Rates on Indicators in SFT Post-Training

The baseline in Figure 4-7 above is Qwen2-0.5B-Instruct. It can be seen that different learning rates have a significant impact on the indicators. A higher learning rate may result in more comprehensive learning, but it may affect the general ability of a certain dimension.

# 5   Conclusion

In this paper, we propose a construction method of continuous post-training alignment data for small language models, which is completely based on open source instruction alignment data. By using the instruction alignment data constructed with our method, we conducted several groups of post-training optimization experiments, including SFT (Supervised Fine Tuning) post-training experiment and KTO (Kahneman Tversky optimization) post-training experiment, as well as SFT-KTO two-stage post-training experiment. At the same time, we also analyzed the impact of weight fusion on model performance, and the impact of different learning rates on model performance. Finally, through the analysis of the general benchmark evaluation indicators, we verified that our method can further improve the performance of the small language models.